\documentclass{article}

\usepackage{PRIMEarxiv}

\usepackage[utf8]{inputenc} 
\usepackage[T1]{fontenc}    
\usepackage{hyperref}       
\usepackage{url}            
\usepackage{booktabs}       
\usepackage{amsfonts}       
\usepackage{nicefrac}       
\usepackage{microtype}      
\usepackage{lipsum}
\usepackage{fancyhdr}       
\usepackage{graphicx} 
\usepackage{float}
\graphicspath{{media/}}     
\usepackage{tabularx}
\usepackage{array}

\usepackage{amsmath}
\usepackage{amssymb}
\usepackage{amsbsy}
\usepackage{amsthm}
\usepackage{algorithm} 
\usepackage{algpseudocode}
\usepackage{booktabs}
\usepackage{siunitx}
\usepackage{natbib}
\usepackage{color}
\title{Bilevel Optimization of Agent Skills via Monte Carlo Tree Search
}

\author{
    Chenyi Huang \\
  Department of Mathematics \\
  National University of Singapore \\
  Singapore\\
  \texttt{e1349254@u.nus.edu} \\
   \And
  Haoting Zhang, Jingxu Xu, Zeyu Zheng\\
  Department of Industrial Engineering and Operations Research\\
  University of California at Berkeley\\
  CA, USA\\
  \texttt{\{haoting\_zhang, jingxu\_xu, zyzheng\}@berkeley.edu} \\
  \And
  Yunduan Lin\\
  Department of Decision, Operation and Technology\\
  The Chinese University of Hong Kong\\
  CHINA\\
  \texttt{yunduanlin@edu.cuhk.hk}
}

\begin{document}
\maketitle

\begin{abstract}
Agent \texttt{skills} are structured collections of instructions, tools, and supporting resources that help large language model (LLM) agents perform particular classes of tasks. Empirical evidence shows that the design of \texttt{skills} can materially affect agent task performance, yet systematically optimizing \texttt{skills} remains challenging. Since a \texttt{skill} comprises instructions, tools, and supporting resources in a structured way, optimizing it requires jointly determining both the structure of these components and the content each component contains. This gives rise to a complex decision space with strong interdependence across structure and components. We therefore represent these two coupled decisions as \texttt{skill} structure and component content, and formulate \texttt{skill} optimization as a bilevel optimization problem. We propose a bilevel optimization framework in which an outer loop employs Monte Carlo Tree Search to determine the \texttt{skill} structure, while an inner loop refines the component content within the structure selected by the outer loop. In both loops, we employ LLMs to assist the optimization procedure. We evaluate the proposed framework on an open-source Operations Research Question Answering dataset, and the experimental results suggest that the bilevel optimization framework improves performance of the agents with the optimized \texttt{skill}.

\end{abstract}

\keywords{LLM Agents, Agent Skills, Skill Optimization, Bilevel Optimization, Monte Carlo Tree Search}

\section{Introduction}
\label{sec:intro}
Large language model (LLM) agents are increasingly deployed to perform complex, multi-step tasks across diverse domains, such as code generation, data processing, and business analysis. A recent trend that enables task specialization in these systems is the agent \texttt{skill} \cite{anthropic2025agentskills}: a collection of software artifacts including task instructions, tools, and supporting resources, organized in a structured folder; see Figure \ref{fig:structure} for an example. During task execution, LLM agents equipped with a particular \texttt{skill} will query and load the components within the \texttt{skill} folder to help task completion. A well-designed \texttt{skill} provides relevant guidance and supporting resources in a compact and usable form, whereas a poorly designed \texttt{skill} may waste context, introduce irrelevant information, misdirect the agent, and in some cases even degrade task performance. Recent benchmark evidence \cite{li2026skillsbench} shows that \texttt{skills} can substantially improve performance, but their effects are highly heterogeneous across tasks. These observations suggest the need to develop systematic strategies for improving \texttt{skills} to optimize agent performance.

Studying this design problem requires a perspective different from what is typically optimized in conventional software artifacts. Conventional software artifacts deliver value mainly through executable code, of which the behavior is largely explicit once implemented. By contrast, a \texttt{skill} affects agent behavior through natural language instructions and auxiliary materials that must be loaded into the context window and interpreted during reasoning. This creates three central challenges for optimization. First, a \texttt{skill} is heterogeneous, combining instructions, executable scripts, reference documents, and structured assets. This enlarges the design space and makes it difficult to represent the optimization problem with an explicit set of decision variables. Second, these components are interdependent. Editing a single component can alter the role, necessity, or interpretation of others, and therefore, each edit of the component requires coordinated revisions elsewhere. Third, even when the decision variables are explicitly represented, the feasible set is discrete and combinatorial, constrained by schema requirements, token-budget limits, and structural consistency conditions. Because of these reasons, \texttt{skill} optimization becomes challenging not only to solve, but even to formulate as a well-structured optimization problem.

In this work, we propose a framework to formulate and solve the \texttt{skill} optimization as a bi-level optimization problem. Our framework is grounded in the official Agent Skills specification (\url{https://agentskills.io/specification}). This specification defines both the structure of a \texttt{skill} and how its contents are exposed during execution. Structurally, a \texttt{skill} is implemented as a directory centered around a required \texttt{SKILL.md} file as the general description, together with optional subdirectories such as \texttt{scripts/} for executable code, \texttt{references/} for supplementary documentation, and \texttt{assets/} for static resources such as templates and lookup tables. The \texttt{SKILL.md} file itself contains two parts: a YAML frontmatter block with structured metadata, such as the skill name, routing description, compatibility requirements, and approved tools, followed by a Markdown body containing the agent-facing instructions. Operationally, the specification adopts a progressive-disclosure loading model over this organization: at agent startup, only lightweight frontmatter metadata is loaded; upon skill activation, the full \texttt{SKILL.md} content is loaded; and files in the optional subdirectories are loaded on demand. This hierarchy motivates us to represent a \texttt{skill} as a tuple $S=(\theta,\phi)$, where $\theta$ denotes the structure configuration of the \texttt{skill}, and $\phi$ denotes the instantiated content within the structure. 

Given this representation, we formulate \texttt{skill} optimization as a bilevel problem in which an outer loop searches over structure configurations $\theta$ and an inner loop optimizes the associated content $\phi$ under each fixed structure. The outer-loop problem is naturally sequential and discrete: structural edits are path-dependent, since an early revision can change which later revisions become feasible, desirable, or necessary. Moreover, the quality of a structure choice is only revealed after subsequent content refinement and downstream evaluation. We therefore organize the revision process as a tree and employ Monte Carlo Tree Search (MCTS) to navigate it. MCTS uses delayed evaluation feedback to favor edit paths that lead to strong empirical performance while preserving exploration of alternative revisions. For each proposed structure, the inner loop first constructs aligned initial content under that structure and then refines it through a bounded sequence of attempts. This refinement is not handled by a single generic procedure. Instead, the inner loop dispatches the proposed structure edit to a refinement family matched to the corresponding type of content work. Refinement then proceeds sequentially within the selected family, i.e., each inner loop explicitly depends on the structure selected by the upstream outer loop. After the allowed attempts are completed, the recorded outcomes are ranked using a conservative selection rule. The top-ranked outcome, together with the evaluation signal, is then returned to the outer loop. In this manner, the bilevel design separates structure search from content refinement and provides an explicit attribution of the evaluation feedback during the optimization procedure. Since both the outer structure search and the inner content refinement require optimization over unstructured variables that are difficult to represent numerically, we employ LLMs to implement the optimization procedures \cite{yang2023large}.

\section{Background and Related Work}
\label{sec:background}
This section reviews two lines of work most relevant to our framework: tree-search methods for LLM agents, and prior work on agent \texttt{skills} as reusable capability modules.

\subsection{Tree Search Methods for LLM Reasoning and Agent Optimization}

Tree-structured search has emerged as a powerful paradigm for improving LLM reasoning and agent decision-making. Yao et al. \cite{yao2023tree} proposed Tree of Thoughts, which extends chain-of-thought prompting by enabling deliberate exploration of multiple reasoning paths over coherent text units (``thoughts''), using search strategies such as breadth-first search and depth-first search for lookahead and backtracking. Building on this, Hao et al. \cite{hao2023rap} introduced Reasoning via Planning, which repurposes the LLM as both a world model and reasoning agent within an MCTS framework. Zhou et al. \cite{zhou2024lats} proposed Language Agent Tree Search, the first general framework that unifies LLM reasoning, acting, and planning through Monte Carlo Tree Search, using LLM-powered value functions and self-reflections together with environment feedback to guide exploration. In the domain of mathematical reasoning, MCTSr \cite{zhang2024mctsr} integrates MCTS with iterative self-refinement and self-evaluation, while rStar-Math \cite{guan2025rstarmath} demonstrates that small language models can match frontier-model performance on competition mathematics through MCTS-based deep thinking with process reward models. Taken together, these works show that tree-based search can improve reasoning by allowing the model to explore, evaluate, and revise intermediate candidates. A natural next step is to apply the same idea not only to reasoning traces, but also to agent artifacts themselves.

Most relevant to our work is AFlow \cite{zhang2025aflow}, which reformulates agentic workflow optimization as a search problem over code-represented workflows and uses MCTS to iteratively refine them through code modification, tree-structured experience, and execution feedback. AFlow demonstrated that MCTS can effectively navigate the discrete, combinatorial space of agent workflow designs, achieving consistent improvements over manually designed baselines. Our framework shares AFlow's use of MCTS for optimizing agent-facing artifacts, but differs in several respects. First, we optimize \texttt{skill} packages rather than code-represented workflows, which introduces a heterogeneous design space spanning instructions, scripts, references, and assets. Second, we adopt a bilevel formulation that explicitly separates structure search from content refinement, whereas AFlow operates on a single level workflow representation. Third, our inner loop uses action-family-aware refinement strategies with a conservative selection rule, rather than applying a uniform code-modification procedure.

Overall, this line of work supports our use of tree search, but existing approaches do not directly address the optimization of structured \texttt{skill} packages. Moreover, because the evaluation of LLM-generated artifacts is inherently noisy, our framework is also related to the broader simulation optimization literature, where ranking-and-selection ideas are used to compare noisy alternatives in discrete design spaces \cite{hong2021surrogate,eckman2021fixed,du2024contextual,fu2026stochastic}.

\subsection{LLM Agent Skills}

Related work on agent \texttt{skills} can be broadly organized into two complementary strands: one studies how reusable capabilities can improve agent behavior, and the other treats \texttt{skills} as portable system components and focuses on their specification, orchestration, evaluation, and security.

Voyager \cite{wang2023voyager} introduced an ever-growing \texttt{skill} library of executable code for a lifelong learning agent in Minecraft, where \texttt{skills} are automatically generated, verified, and stored for future retrieval. Reflexion \cite{shinn2023reflexion} proposed verbal reinforcement learning, in which agents improve through linguistic self-reflection rather than weight updates, storing experiential feedback in an episodic memory. More recently, SAGE \cite{wang2025sage} uses reinforcement learning to enable agents to build and refine reusable \texttt{skill} libraries, demonstrating self-improvement through iterative \texttt{skill} acquisition. These works establish the principle that an agent's capabilities can be improved by optimizing a library of reusable modules rather than the underlying model.

In parallel, a growing body of work has treated agent skills as first-class software artifacts. Anthropic introduced the Agent \texttt{Skills} specification \cite{anthropic2025agentskills}, which defines a structured directory format centered on a \texttt{SKILL.md} file with progressive-disclosure loading. This specification has since been adopted as an open standard for cross-platform \texttt{skill} portability. Li et al. \cite{li2026agentskillos} proposed AgentSkillOS, an operating-system-like framework for organizing and orchestrating skills at the ecosystem scale. On the evaluation side, SkillsBench \cite{li2026skillsbench} provides a large-scale benchmark for assessing how well agent skills generalize across diverse tasks, revealing that \texttt{skill} effects are highly heterogeneous and that \texttt{skill} quality significantly impacts downstream performance. Complementary security analyses \cite{liu2026agentskillswild} have documented the prevalence of vulnerabilities and malicious patterns in deployed \texttt{skills}, underscoring the importance of rigorous skill validation.

\section{Methodology}
\label{sec:methodology}

\subsection{Framework Overview}
\label{subsec:framework_overview}

We now present the overall framework for \texttt{skill} optimization. To study this problem systematically, we represent a \texttt{skill} as a tuple $S=(\theta,\phi)$. The variable $\theta\in\Theta$ denotes the structure configuration of the \texttt{skill}, that is, the set of components it contains and their organizational arrangement. The variable $\phi\in\Phi(\theta)$ denotes the instantiated content under that structure, including the instructions in \texttt{SKILL.md}, the code in executable scripts, the text in reference documents, and the data stored in static assets. The structure space $\Theta$ is discrete and combinatorial because it ranges over alternative structure configurations. In contrast, for a fixed structure $\theta$, the space $\Phi(\theta)$ consists of all content instances compatible with that structure. This space is typically extremely large, since the instructions, code, and supporting materials instantiated under the given structure may still vary widely.

\begin{figure}[htb]
    \centering
    \includegraphics[width=\linewidth]{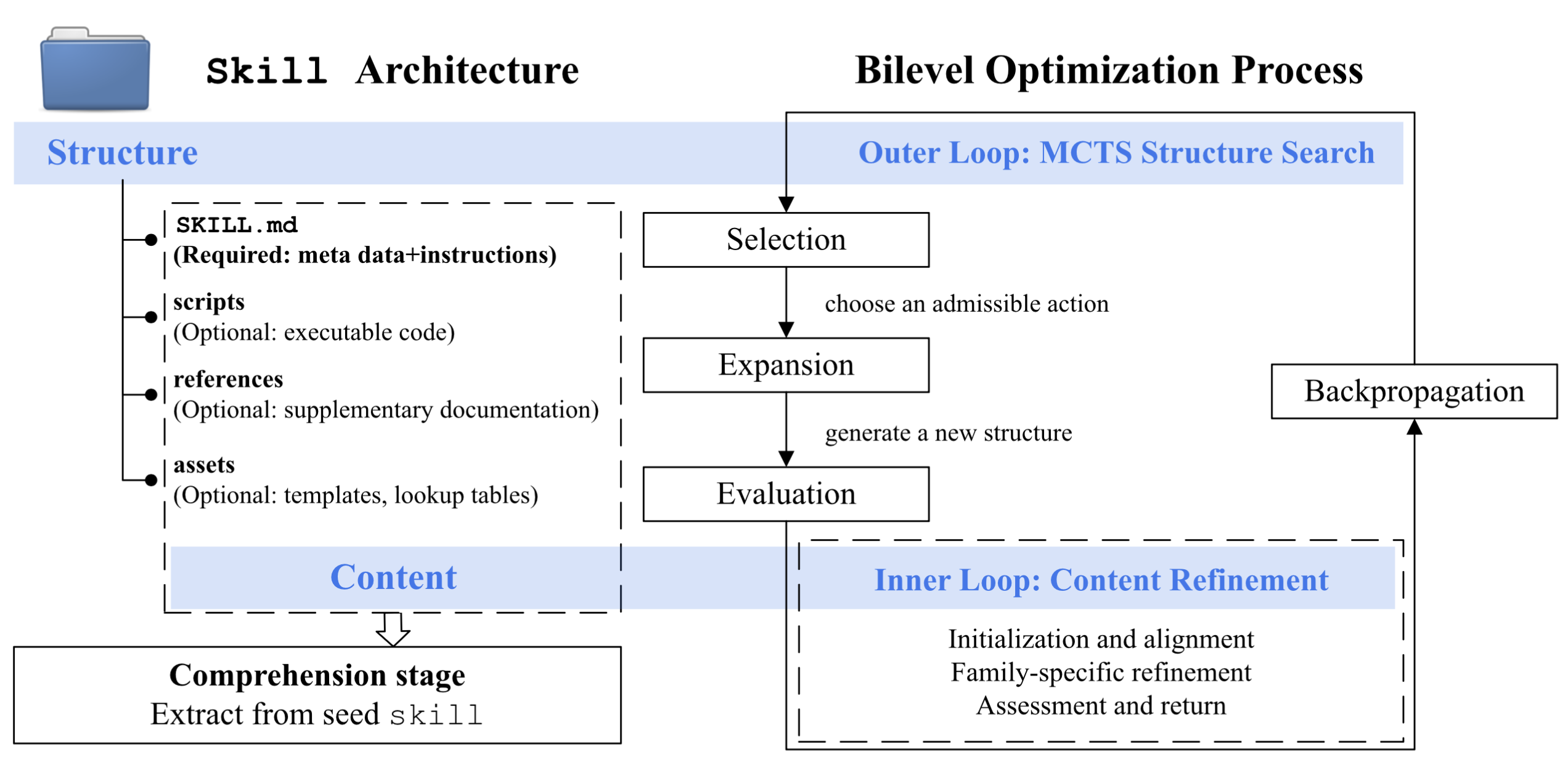}
    \caption{Bilevel Optimization Architecture for Agent Skill Modules}
    \label{fig:structure}
\end{figure}

For a given seed \texttt{skill} $S_0$, we formulate \texttt{skill} optimization as the bilevel problem
\[
\max_{\theta \in \Theta}\;\max_{\phi \in \Phi(\theta)}\; R_{S_0}(\theta,\phi),
\]
subject to structural validity and token budget constraints. The objective $R_{S_0}(\theta,\phi)$ is the scalar performance score assigned to candidate $(\theta,\phi)$ by the evaluation procedure associated with $S_0$. It is seed-dependent because the downstream tasks and assessment criteria depend on the intended function of the seed \texttt{skill}. Structure-validity constraints require that a candidate remain a well-formed \texttt{skill} under the specification, for example, by preserving the required directory structure and maintaining a well-formed \texttt{SKILL.md}. Token-budget constraints require the candidate to remain usable within the context window, for example, by preventing the active instructions from becoming too long or the overall directory contents from becoming excessively large. This bilevel formulation reflects the hierarchical nature of the problem: the outer level compares candidate structures, while the inner level evaluates each structure through the best content refinement achievable within it, as illustrated in Figure~\ref{fig:structure}.

To address this problem, we develop a framework that begins with a one-time comprehension stage, which converts a raw seed \texttt{skill} into an optimization-ready initialization, and then improves the \texttt{skill} through bilevel optimization with an outer loop over structure candidates and an inner loop over content refinement.

The \textbf{comprehension stage} is a one-time preparation step performed before bilevel optimization begins. Applied to the seed \texttt{skill} $S_0$, it first extracts the files in the directory, infers their structural roles, and parses the frontmatter and section organization of \texttt{SKILL.md}, yielding the initial structure representation $\theta_0$. Using $\theta_0$ together with the extracted file contents, it then instantiates the initial aligned content representation $\phi_0$. In addition, the comprehension stage constructs a \texttt{skill} profile $\mathcal{P}$, which summarizes the task the \texttt{skill} is intended to support, the relevant success criteria and quality dimensions, and the structural revision directions that appear most promising. We use $\mathcal{P}$ to define a profile-guided search region $\widetilde{\Theta}(\mathcal{P}) \subseteq \Theta$. In this way, $\mathcal{P}$ acts as an algorithmic search prior for the outer loop: it narrows the structure search space, guides how structure candidates are explored and provides persistent task-aligned context for exploring candidate revisions. Together, $(\theta_0,\phi_0)$ and $\mathcal{P}$ provide a grounded initialization for the subsequent optimization.

The \textbf{bilevel optimization} then starts with the outer level, which searches over alternative structure candidates starting from $\theta_0$. We model this process as a sequential search over a tree of candidate structures, where each node corresponds to a structure state and each edge corresponds to an admissible edit. Because candidate structures are generated, evaluated, and updated iteratively, the search is carried out through an outer loop. The tree-based view reflects the path-dependent nature of structure optimization: an earlier edit can change which later edits become desirable, feasible, or even necessary. For example, adding a detailed instruction section to \texttt{SKILL.md} may later require moving supporting details into a separate reference file to remain within the token budget. Guided by the prior $\mathcal{P}$, the outer loop repeatedly selects a promising structure state, proposes a candidate edit, and validates the resulting candidate against structural and budget constraints. Only feasible candidates are passed to the inner level for content refinement and downstream evaluation. The resulting reward and diagnostics are then returned to the outer loop to update the search statistics and guide subsequent exploration. We implement the search at this level using Monte Carlo Tree Search (MCTS), which is well suited to discrete sequential optimization with delayed feedback, since the value of a structure revision is revealed only after content refinement and evaluation at the inner level.

Once the outer loop generates a valid structure candidate, the inner level refines the associated content while keeping the selected structure fixed. Because the outer loop modifies the structure, the system must first transfer the existing content into the new structure through a bridge operation that preserves reusable material while accommodating newly introduced components. Starting from this aligned content state, refinement is carried out through an inner loop. This inner loop performs a bounded number of refinement attempts, with the specific refinement procedure determined by the type of structural edit that triggered it. Each valid refinement outcome is evaluated on the downstream tasks under the current structure, producing a corresponding reward and diagnostic record. These evaluated outcomes are then compared, and one final content is selected for return to the outer loop. Combined with the fixed structure, this selected content determines the candidate \texttt{skill} associated with the current structure change.

After multiple rounds of outer-loop search and inner-level refinement, the framework returns the \texttt{skill} candidate associated with the highest-reward node in the search tree.

\subsection{Outer Loop: MCTS-Based Structure Search}
\label{subsec:outer_loop}
In this subsection, we describe the main components and procedure of the MCTS-based outer loop.

We use $n$ to denote a node in the search tree and $\mathcal{V}$ to denote the current set of tree nodes. The root corresponds to the initial candidate $(\theta_0,\phi_0)$. Each node $n$ stores a structure $\theta$, its associated aligned content state $\phi$, and search statistics including the visit count $N(n)$ and the current estimate of its search value $\bar{Q}(n)$. Although both $\theta$ and $\phi$ are stored at each node, the action space of the outer loop operates only on the structure. The content $\phi$ is retained so that, after a structural revision is proposed, the resulting candidate can be refined by the inner loop and evaluated consistently. The action space $\mathcal{A}$ consists of admissible structure edits, including additions, removals, reorderings, and edits to components such as sections, references, scripts, assets, and metadata. The set of admissible actions varies across nodes, because whether a structural edit can be applied depends on the current structure. Given a current structure $\theta$ and an admissible action $a \in \mathcal{A}$, a transition map $T$ defines the successor structure $\theta' = T(\theta, a)$.

The outer loop follows the four-step structure of MCTS: \textbf{selection}, \textbf{expansion}, \textbf{evaluation}, and \textbf{backpropagation}. In each iteration, it selects a parent node, proposes and applies a structure edit, evaluates the resulting candidate after inner-loop refinement, and backpropagates the reward to update the tree statistics. In our setting, this basic procedure is adapted in two ways. First, evaluation is bilevel: a structure candidate is evaluated only after its content has been refined by the inner loop and assessed on downstream tasks. Second, the search is LLM-guided: LLM reasoning is used to analyze the current state, interpret evaluation feedback, and propose the next structural revision.

The MCTS steps begin with \textbf{selection}, which chooses an existing node in the current search tree as the parent for the next expansion. The framework supports two selection policies. The default policy is upper confidence bound (UCB), which assigns each selectable node $n \in \mathcal{N}$ the score
\[
\mathrm{UCB1}(n)=\bar{Q}(n)+c\sqrt{\frac{\ln \sum_{m\in\mathcal{V}}N(m)}{N(n)}},
\]
where $\bar{Q}(n)$ is the empirical mean reward and $c > 0$ is the exploration constant. The framework selects node $n^\star = \arg\max_{n \in \mathcal{N}} \mathrm{UCB1}(n)$. This policy is suitable when the reward signal is relatively stable.

For noisier evaluation settings, the framework also supports a mixed-probability policy that combines uniform exploration with score-based sampling. For each node $n \in \mathcal{N}$, define the centered value
$
\tilde{Q}(n)=\bar{Q}(n)-\frac{1}{|\mathcal{N}|}\sum_{m\in\mathcal{N}}\bar{Q}(m).
$
The sampling distribution is
\[
P(n)=\lambda \cdot \frac{1}{|\mathcal{N}|}+(1-\lambda)\cdot
\frac{\exp\!\bigl(\alpha \tilde{Q}(n)\bigr)}
{\sum_{m\in\mathcal{N}}\exp\!\bigl(\alpha \tilde{Q}(m)\bigr)},
\qquad n\in\mathcal{N},
\]
where $\lambda \in [0,1]$ is the mixing coefficient and $\alpha > 0$ is an inverse-temperature parameter. When $\lambda$ is large, the policy approaches uniform exploration; when $\lambda$ is small, it places great weight on high-value nodes. This provides a smooth interpolation between broad exploration and focused exploitation, mitigating the risk of premature convergence when reward estimates are highly stochastic.

After a node has been selected, the framework proceeds to \textbf{expansion}, which generates a new structure candidate from the selected parent node through a three-stage LLM-guided reasoning procedure. This design is motivated by evidence that decomposing complex tasks into intermediate steps can improve performance \cite{wei2022chain,zhou2022least}. In the \textit{analysis} stage, the system examines the current \texttt{skill} structure together with summary evaluation information, the persistent \texttt{skill} profile, and specification constraints, aiming to form a structural understanding of the current candidate. In the \textit{diagnosis} stage, this structural understanding is combined with richer evaluation diagnostics and search experience from previous rounds to identify the likely root cause of underperformance and a structural hypothesis for addressing it. In the \textit{proposal} stage, the diagnosis is translated into a concrete action $a \in \mathcal{A}$, conditioned on the available action types, their parameter definitions, and warnings from previous failed or risky edits. 

Once an action $a$ is proposed, it is applied to produce $\theta' = T(\theta, a)$. Before $\theta'$ is admitted into the tree, the framework applies a structural validation gate to check whether the revised candidate remains a valid \texttt{skill} and satisfies the required structural and budget constraints. Candidates that fail these checks are discarded, so that only feasible successors proceed to content refinement.

The next phase is \textbf{evaluation}. Because the outer loop modifies only the structure, a newly proposed $\theta'$ cannot be evaluated directly. It must first be passed to the inner loop. Under this fixed structure $\theta'$, the inner loop constructs aligned content, refines it, and selects one final refined content $\phi^\star(\theta')$. The pair $(\theta',\phi^\star(\theta'))$ then defines the candidate \texttt{skill} for the current expansion. This candidate is evaluated on the downstream tasks, and its resulting task performance defines the reward $R_{S_0}(\theta',\phi^\star(\theta'))$ together with the associated diagnostics. The evaluation also produces diagnostic information, which is returned together with the reward for use in subsequent search steps. The details of the inner-loop refinement and evaluation procedure are given in Section~\ref{subsec:inner_loop}.

The framework then proceeds to \textbf{backpropagation}. The reward $R_{S_0}(\theta', \phi^\star(\theta'))$ is propagated along the path from the expanded node back to the root, updating the visit counts and value estimates of the visited nodes. Through these updates, the search tree gradually records which structural revisions tend to yield stronger downstream performance after content refinement.

The four phases together complete one iteration of the outer loop. The search terminates when the iteration budget is exhausted, when no valid expansions remain, or when anti-stagnation controls indicate that further exploration is unlikely to improve performance. 

\subsection{Inner Loop: Content Refinement}
\label{subsec:inner_loop}

We now turn to the inner loop, which is invoked during the evaluation phase of the outer loop. Once the outer loop proposes a structurally valid candidate $\theta'$, the inner loop refines the associated content under that fixed structure. Because the content may include instruction text, reference material, and executable code, this refinement is carried out under a bounded optimization budget and proceeds in three phases: \textbf{initialization and alignment}, \textbf{family-specific refinement}, and \textbf{pessimistic assessment and return}.

The first phase, \textbf{initialization and alignment}, introduces the bridge step that connects outer-loop structure search with inner-loop content refinement. When the outer loop proposes a new structure $\theta'$, it changes only the organization of the \texttt{skill}, rather than the concrete content instantiated under that organization. The candidate is therefore not yet directly executable. To make refinement possible, the framework first constructs aligned initial content $\phi_0(\theta')$ by transferring existing content into the new structure organization, preserving reusable material whenever possible while accommodating any added, removed, or reordered components. This resulting $\phi_0(\theta')$ then provides the initial state for subsequent content refinement.

The second phase is \textbf{family-specific refinement.} Different structure changes induce different content-level optimization needs,  so the inner loop does not apply a single generic refinement procedure. Instead, each proposed change is dispatched to a refinement family matched to the kind of content work it requires. The framework implements five refinement families: metadata-light updates, metadata-routing text, instruction text, redistribution, and script edit or generation. For example, a change to section organization in \texttt{SKILL.md} naturally leads to instruction text refinement, whereas a script-related edit leads to script editing or generation. Once the refinement family has been selected, the inner loop performs a bounded sequence of $M$ refinement attempts within that family. The attempt is the basic sequential unit of refinement. Starting from the aligned content $\phi_0(\theta')$, attempt $m$ takes the current content package as input and returns one refined package as output, which then becomes the input to attempt $m+1$. This produces the trajectory
$
\phi_0(\theta') \;\rightarrow\; \phi^{(1)} \;\rightarrow\; \cdots \;\rightarrow\; \phi^{(M)},
$
where $\phi^{(m)}$ denotes the content returned by attempt $m$. The process terminates when the budget is exhausted or when further refinement is judged unnecessary.

Each attempt produces an outcome record containing the returned content together with evaluation signals such as its improvement over a comparison baseline, and a confidence measure summarizing the strength of the supporting evidence. In some refinement families, a single attempt may internally compare multiple candidate variants before returning one selected content. When this occurs, and the candidate variants yield improvements $\delta_1, \dots, \delta_k$ relative to the comparison baseline, the inner loop computes a lower confidence bound
$
\mathrm{LCB} = \bar{\delta} - t_{\mathrm{crit}}\,{s}/{\sqrt{k}},
$
where $\bar{\delta}$ is the mean improvement, $s$ is the sample standard deviation, and $t_{\mathrm{crit}}$ is a conservative critical value. This quantity discounts the observed mean improvement by an uncertainty penalty and serves as a pessimistic estimate of gain reliability.

The third phase, \textbf{assessment and return}, determines which refinement outcome is passed back to the outer loop. Since downstream evaluation is inherently noisy, a positive observed gain does not by itself guarantee a reliable refinement. The framework therefore uses the lower confidence bound from the second phase as a conservative acceptance criterion: an outcome is treated as having passed the pessimistic gate when its estimated lower confidence bound is nonnegative, that is, when $\mathrm{LCB}\ge 0$. The recorded outcomes are then ranked by first preferring those that pass this pessimistic gate, then those with larger improvement, and finally those with higher confidence. Under this ordering, the top-ranked content package is selected as the accepted content. The accepted content and its diagnostics are returned to the outer loop, which uses the resulting reward for backpropagation. The complete procedure of the \texttt{skill} optimization is summarized in Algorithm~\ref{alg:bilevel_mcts}.

\begin{algorithm}[htb]
\caption{Bilevel Skill Optimization via MCTS}
\label{alg:bilevel_mcts}
\begin{algorithmic}[1]
\Require Seed skill $S_0$, Maximum search rounds $K$
\Ensure Optimized skill $S^\star = (\theta^\star, \phi^\star)$

\State Extract initial structure $\theta_0$, content $\phi_0$, and task profile $\mathcal{P}$ from $S_0$
\State Initialize search tree $\mathcal{T}$ with root node $n_0 = (\theta_0, \phi_0)$
\State Initialize best candidate: $\theta^\star \gets \theta_0$, $\phi^\star \gets \phi_0$, and best reward $r^\star \gets -\infty$

\For{$t=1, 2, \dots, K$}
    \State Select a parent node $n \in \mathcal{T}$ containing structure $\theta_n$ and content $\phi_n$
    \State Propose structure-editing action $a$ via LLM conditioned on $\theta_n$ and $\mathcal{P}$
    
    \If{action $a$ is structurally valid and meets budget constraints}
        \State $\theta' \gets \text{ApplyAction}(\theta_n, a)$ \Comment{Outer loop: create new structure}
        
        \State $\phi'_0 \gets \text{AlignContent}(\phi_n, \theta', a)$ \Comment{Inner loop: bridge existing content}
        \State $\phi' \gets \text{RefineContent}(\phi'_0, \theta', a)$ \Comment{Inner loop: perform bounded refinement}
        
        \State $r' \gets R_{S_0}(\theta', \phi')$ \Comment{Evaluate downstream task performance}
        
        \State Add new node $n' = (\theta', \phi')$ to $\mathcal{T}$ and backpropagate $r'$
        
        \If{$r' > r^\star$}
            \State $\theta^\star \gets \theta'$; $\phi^\star \gets \phi'$; $r^\star \gets r'$ \Comment{Update optimal reward}
        \EndIf
    \EndIf

    \If{search convergence criteria are met}
        \State \textbf{break} \Comment{Terminate early if search stagnates}
    \EndIf
\EndFor

\State \Return $S^\star = (\theta^\star, \phi^\star)$
\end{algorithmic}
\end{algorithm}

\section{Experiments}
\label{sec:experiments}

In this section, we evaluate the proposed method by optimizing a seed \texttt{skill} for the Operations Research Question Answering (ORQA) benchmark dataset \cite{mostajabdaveh2025evaluating}. ORQA is a multiple choice question answering task in operations research, in which the agent receives a natural language problem description together with answer options and must infer the underlying optimization model in order to choose the correct answer. Typical questions involve identifying decision variables, constraints, parameters, objectives, or the meaning of modeled expressions. The full ORQA benchmark contains 1,513 instances across 20 application domains, with an average input length of 231 words. Table~\ref{tab:orqa_examples} presents two representative ORQA examples. 
\begin{table*}[htb]
\centering
\caption{Representative ORQA examples.}
\label{tab:orqa_examples}
\small
\renewcommand{\arraystretch}{1.35}
\begin{tabularx}{\textwidth}{>{\raggedright\arraybackslash}X >{\raggedright\arraybackslash}p{0.17\textwidth} >{\raggedright\arraybackslash}p{0.32\textwidth} c}
\toprule
\textbf{Context (abridged)} & \textbf{Question} & \textbf{Options} & \textbf{Ans.} \\
\midrule
A broadcast network must choose which shows to air on a single channel. Each show has a duration, a deadline, and a viewer-rating score. Some shows may be omitted if they conflict with others. The goal is to maximize total viewer rating while respecting scheduling constraints. &
What are the decision activities of the optimization problem? &
(A) Due date \newline
(B) Show broadcast order \newline
(C) Show broadcast indicator \newline
(D) Processing time &
(C) \\
\midrule
A logistics manager must plan the movement of trucks so that all deliveries are completed using as few vehicles as possible. The data include locations, travel times, delivery requirements, and initial vehicle availability at each location. &
Which of the following options are participating decision activities in the objective criterion for this problem? &
(A) Number of trucks departing daily from each location \newline
(B) Number of deliveries to be made from each location \newline
(C) Whether to set a location to never be an origin \newline
(D) Travel time between each location &
(A) \\
\bottomrule
\end{tabularx}
\end{table*}

The seed \texttt{skill} used in the ORQA experiment is a small two-file \texttt{skill} package generated by \texttt{skill-creator}, a built-in \texttt{skill} for creating and modifying other \texttt{skills}. Our aim is to further optimize this AI-generated seed \texttt{skill} for answering ORQA questions. It consists of a main \texttt{SKILL.md} file and a supporting reference file on question types. The main \texttt{SKILL.md} file provides the agent a basic answering workflow: restate the optimization goal, identify the type of model component being queried, rule out clearly incompatible options, relate the remaining choices back to the problem description, and return exactly one final answer token. It also provides a few heuristics and final checks. The supporting reference file summarizes common ORQA question types and the distinctions among objectives, constraints, variables, parameters, and indexed sets. This seed \texttt{skill} serves both as the starting point for optimization and as the baseline for comparison. During optimization, candidate \texttt{skills} must satisfy the structural-validity constraints of the Agent Skills specification and remain within the size budget. In particular, the active \texttt{SKILL.md} content is constrained by an activation budget of 5000 tokens, with a warning threshold at 3500 tokens; reference files are loaded only on demand, and script code itself is not directly loaded into the model context.



\subsection{Experiment Setup}

In this subsection, we describe the experiment procedure and parameter settings used to evaluate the proposed bilevel optimization framework on ORQA.

To support bilevel optimization, we partition the original ORQA dataset into \textit{search}, \textit{confirm}, and \textit{test}, whose roles parallel the standard training, validation, and test sets commonly used in machine learning literature. The \textit{search} split is used during optimization to evaluate candidate \texttt{skills} and guide the search process. The \textit{confirm} split is used after optimization for model selection across alternative search settings. Finally, the \textit{test} split is held out until the end and used exclusively for the final evaluation of the selected \texttt{skill}. To keep the optimization cost manageable, each run uses sampled subsets from these splits; in the reference run, the total sample size is 120 questions.

To ensure consistency, we fix the main LLM and agent settings throughout the experiment. The LLM agents operate in two roles: runtime evaluation, where candidate \texttt{skills} are executed in a Harbor sandbox using \texttt{openai/gpt-5.2-codex}; and optimization, where the structure search is orchestrated by \texttt{openai/gpt-5.4} through DSPy. Token limits are allocated strategically across five pipeline stages as per-call maximums. Specifically, the initial comprehension stage uses 1024 tokens. Within the outer loop, the analysis, diagnosis, and proposal stages are capped at 1536, 1024, and 20000 tokens, respectively. In the inner loop, each content-refinement subroutine is restricted to 1024 tokens. The significantly larger budget reserved for the proposal stage accommodates the complex task of synthesizing the current structural state, evaluation feedback, and search history to formulate the next revision. This strategic allocation of varying computational budgets across different pipeline stages shares conceptual similarities with multi-fidelity simulation optimization, where expensive, high-fidelity evaluations are strictly reserved for the most critical decision-making points \cite{rhodes2018multi}.

Based on these settings, we then compare two search configurations, each corresponding to one fixed set of MCTS hyperparameters. Configuration A is the more conservative setting, using fewer search rounds, deterministic UCB1 selection, earlier convergence, and priority-action-whitelist enforcement. Configuration B is the more exploratory setting, using more search rounds, mixed-probability selection, more patient convergence settings, and a less restricted action space. Table~\ref{tab:orqa_sweep_configs} summarizes the main search, convergence, and selection-policy settings. Each configuration produces one optimal candidate \texttt{skill} on the \textit{search} split. These configuration-level candidate \texttt{skills} are then re-evaluated on the \textit{confirm} split, and the final winner is the one with the highest score on the \textit{confirm} split. Ultimately, this selected winner and the initial seed \texttt{skill} are evaluated once on the \textit{test} split to measure final performance improvements.

\begin{table}[htb]
\centering
\caption{Sweep configurations used in the ORQA experiment.}
\label{tab:orqa_sweep_configs}
\begin{tabular}{lcc}
\hline
Setting & Configuration A & Configuration B \\ \hline
Max rounds & 3 & 6 \\
Selection policy & UCB1 & Mixed-probability \\
Exploration constant & 1.2 & -- \\
\(\alpha\) & -- & 0.55 \\
\(\lambda\) & -- & 0.25 \\
Min rounds before convergence & 2 & 3 \\
Consecutive stale rounds to stop & 2 & 3 \\
Improvement threshold & 0.001 & 0.001 \\ \hline
\end{tabular}
\end{table}

\subsection{Results}
\label{subsec:result}

In this subsection, we report the experiment results. 

During the optimization procedure, both Configuration A and Configuration B reached the same peak reward of \(0.9434\) on \textit{search} split. The deciding difference emerged at the confirmation stage: Configuration A achieved a mean score of $0.8571$ on the sampled \textit{confirm} split, whereas Configuration B achieved $0.8857$. Consequently, the winning skill was selected from Configuration B. When evaluated on the held-out \textit{test} split, the seed \texttt{skill} established a baseline exact-match score of $0.90625$. The final optimized \texttt{skill} from Configuration B achieved a score of $0.9375$, yielding an overall improvement of $+0.03125$.

\begin{figure}[htb]
    \centering
    \includegraphics[width=\linewidth]{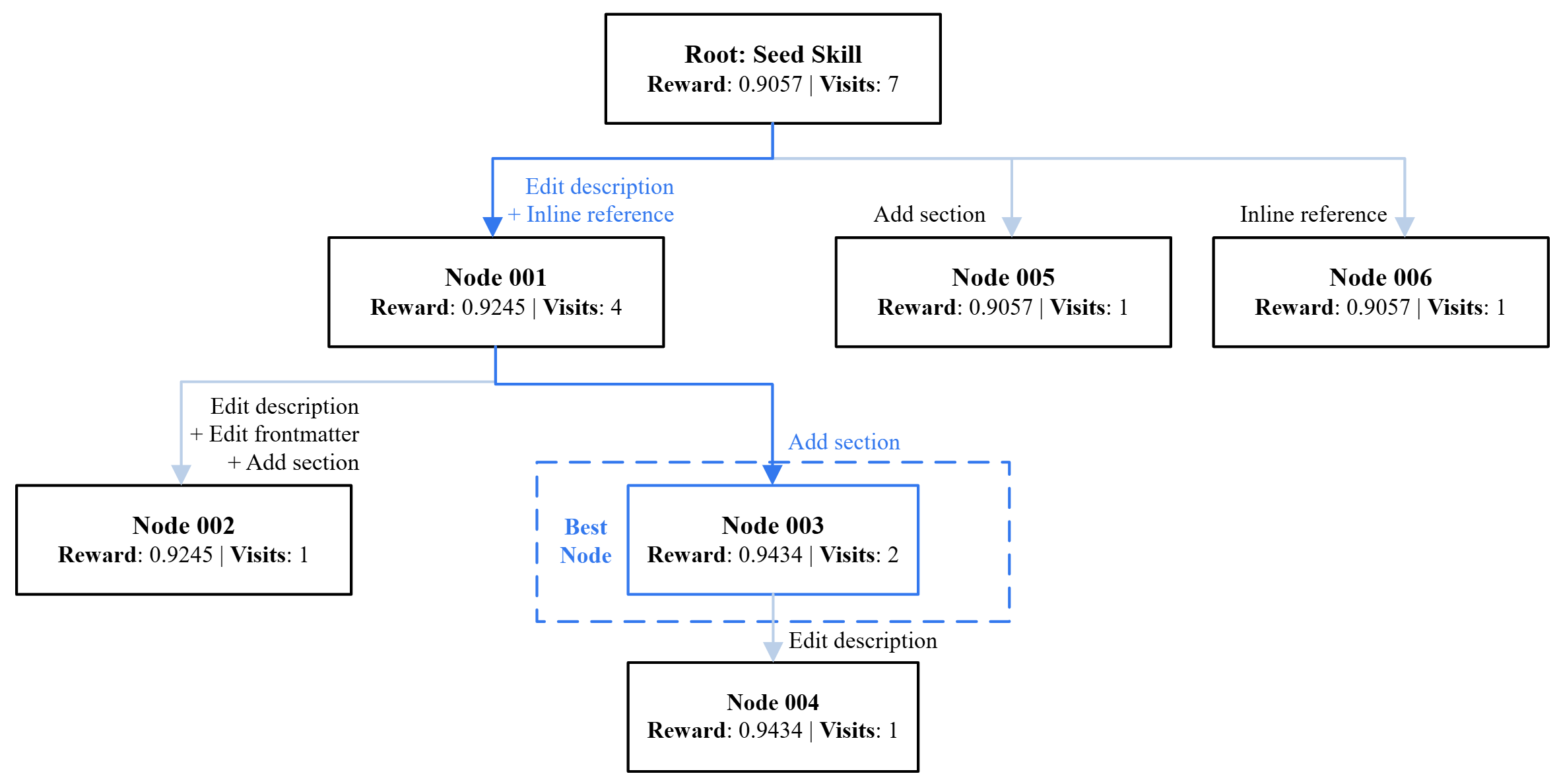}
    \caption{Illustration of MCTS search tree for Configuration B in the ORQA experiment. Nodes are labeled by the structure action and the resulting reward on \textit{search} split. The winning path and selected node are highlighted in blue, while weaker alternatives are shown in a faded style.}
    \label{fig:orqa_mcts_tree}
\end{figure}

Figure~\ref{fig:orqa_mcts_tree} illustrates the actual MCTS search tree produced by Configuration B. The figure clearly delineates the winning path, demonstrating that the outer loop explored several structure alternatives before finalizing the \texttt{skill}. Starting from the seed \texttt{skill}, the search first promoted a composite edit that revised the \texttt{skill} description and relocated the key question-type guidance from the reference file directly into \texttt{SKILL.md}. Subsequently, it added a dedicated \texttt{Question-Type Triage Checklist} section, which yielded the best reward in the tree on \textit{search} split. The figure also displays branches that were explored but ultimately abandoned. This highlights that the search algorithm actively compared multiple structural options rather than defaulting to a single, fixed revision path. 

Next, we provide a comparative analysis of the winning \texttt{skill} and the seed \texttt{skill}, focusing on their concrete differences and the underlying drivers of the observed performance improvements.

At the structure level, the seed skill has two files: a main \texttt{SKILL.md} file and a separate reference file on question types. By contrast, the winning skill keeps only a single \texttt{SKILL.md} file. The key question-type guidance that was previously stored in the separate reference file is moved into the main instruction file, and the winning skill also adds a dedicated \texttt{Question-Type Triage Checklist} section. As a result, the skill becomes more self-contained and easier to follow. This centralization likely drives performance improvements, as the agent can now process critical classification guidance directly within its primary instruction surface rather than navigating distributed files.

At the content level, the winning \texttt{SKILL.md} is far more explicit regarding the agent's expected actions and execution sequence. Its frontmatter description clearly defines the skill as an ORQA model-formulation skill, and its input contract is both expanded and strictly enforced. The newly introduced triage section mandates that the agent identify the question type prior to initiating the main workflow. This ensures the agent distinguishes true model-formulation multiple-choice items from other OR questions before attempting to map variables, constraints, parameters, and objectives. Furthermore, the core workflow is rewritten as a clearer step-by-step procedure, while internal heuristics and hard checks are reinforced to better enforce strict answer-type matching, systematic option elimination, and single-token output. These content refinements enhance task execution by reducing ambiguity, formalizing the reasoning process, and tightening the conditions under which a final answer is selected.

Taken together, the comparison suggests that the overall performance gain stems from synergistic improvements in both structure and content. The outer loop successfully reorganizes the \text{skill}, migrating essential guidance into a highly accessible structural format, while the inner loop contributes by making that guidance clearer, more explicit, and more constrained.

\bibliographystyle{unsrt}  
\bibliography{references}

\end{document}